# Kinematic and Dynamic Analysis of the 2-DOF Spherical Wrist of Orthoglide 5-axis

Raza UR-REHMAN, Stephane CARO, Damien CHABLAT and Philippe WENGER

Institut de Recherche en Communication et Cybernétique de Nantes, UMR CNRS 6597, 1 rue de la Noë, 44321, Nantes Cedex 03, France
Raza.Ur-Rehman@irccyn.ec-nantes.fr
Stephane.Caro@irccyn.ec-nantes.fr
Damien.Chablat@irccyn.ec-nantes.fr
Philippe.Wenger@irccyn.ec-nantes.fr

**Abstract -** This paper deals with the kinematics and dynamics of a two degree of freedom spherical manipulator, the wrist of Orthoglide 5-axis. The latter is a parallel kinematics machine composed of two manipulators: i) the Orthoglide 3-axis; a three-dof translational parallel manipulator that belongs to the family of Delta robots, and ii) the Agile eye; a two-dof parallel spherical wrist. The geometric and inertial parameters used in the model are determined by means of a CAD software. The performance of the spherical wrist is emphasized by means of several test trajectories. The effects of machining and/or cutting forces and the length of the cutting tool on the dynamic performance of the wrist are also analyzed. Finally, a preliminary selection of the motors is proposed from the velocities and torques required by the actuators to carry out the test trajectories.

**Résumé -** L'objet de ce papier est de présenter une étude cinématique et dynamique d'un mécanisme sphérique à deux degrés de liberté (ddls), le poignet de l'Orthoglide 5 axes. Ce dernier est une machine à cinématique parallèle composée de deux manipulateurs: i) l'Orthoglide 3 axes, manipulateur parallèle à trois ddls de translation qui appartient à la famille des robots de type Delta, et ii) l'Oeil Agile; poignet sphérique à deux ddls. Les paramètres géométriques et inertiels sont déterminés à l'aide d'un logiciel de CAO. La performance du poignet sphérique est soulignée par plusieurs trajectoires tests. Les effets de l'effort d'usinage et de la longueur de l'outil sur la performance dynamique du poignet sont également analysés. Enfin, une sélection préliminaire des moteurs est proposée à partir des vitesses et des couples requis par les actionneurs afin d'effectuer les trajectoires désirées

**MOTS-CLÉS:** Mécanisme sphérique/Oeil Agile /Orthoglide 5-axis/cinématique/dynamique

## 1 Introduction

A spherical mechanism can orient or move the end-effector about the center of rotation of the mechanism. A typical 3-DOF spherical manipulator or wrist provides the three dimensional (3D) rotations like a human hand but in most robotics applications only 2D rotations are sufficient. Rotation about the axis of symmetry of the end-effector is not necessary and if required, can be provided independently.

Several researchers have worked in the domain of spherical mechanisms mainly for the applications of end-effector orientations. One of the earlier spherical mechanisms is that presented in the work of Asada and Cro Granito [1], which is a three-degree-of-freedom (dof) spherical wrist with coaxial motors and three kinematics chains. In 1989, W. M. Craver [2], analysed a spherical robotic shoulder module. Other major publications to the research and development of the spherical mechanism are the work of Gosselin et al [3, 4, 5] where optimum kinematic designs of different types of spherical parallel mechanisms are presented. The Agile Eye, one of the most famous spherical mechanisms designed by Gosselin and Hamel [6], is a 3-dof parallel mechanism developed to control the orientation of a camera. Several mechanisms have been designed for diverse applications from the Agile Eye. Spherical mechanisms can be implemented in radar applications [7], camera manipulations [8] and surgical applications [9]. Cavallo and Michelini [10] introduced a 3-dof spherical parallel mechanism composed of three identical kinematic chains to orient the propeller and duct of a small autonomous underwater vehicle (AUV). Main contributions for the design and analysis of spherical mechanisms are reported in [1, 3, 4, 5, 6, 11, 12, 13, 14, 15, 16].

This paper focuses on the kinematic and dynamic analysis of the two dof spherical wrist of Orthoglide 5 - axis, a five dof parallel kinematics machine developed for high speed operations. Here, we focus on the evaluation of the velocities, accelerations and the torques required by the actuators of the spherical wrist. First, the kinematics of the spherical wrist is studied and then its dynamics is analyzed by means of the Newton-Euler approach. The geometric and inertial parameters are determined with a CAD software. The performance of the manipulator is emphasized by means of several test trajectories. Finally, the actuators are selected in the catalogue based on the velocities and torques required by the actuators to carry out the test trajectories.

## 2 Orthoglide 5-axis

Orthoglide 5-axis, illustrated in Figure 1, is derived from a 3-dof translating manipulator, the Orthoglide 3-axis and a 2-dof spherical wrist [17].

Orthoglide 3-axis, is a Delta-type PKM [18] dedicated to 3-axis rapid machining applications developed at the Institut de Recherche en Communications et Cybernétique de Nantes (IRCCyN) [19]. This mechanism is composed of three identical legs. Each leg is made up of a prismatic joint, a revolute joint, a parallelogram joint and another revolute joint. Only the prismatic joints of each leg are actuated.

It gathers the advantages of both serial and parallel kinematic architectures such as regular workspace, homogeneous performances, good dynamic performances and stiffness. The interesting features of Orthoglide 3-axis are a regular dexterous workspace, uniform kinetostatic performances in all directions, good compactness [20] and high stiffness [21].

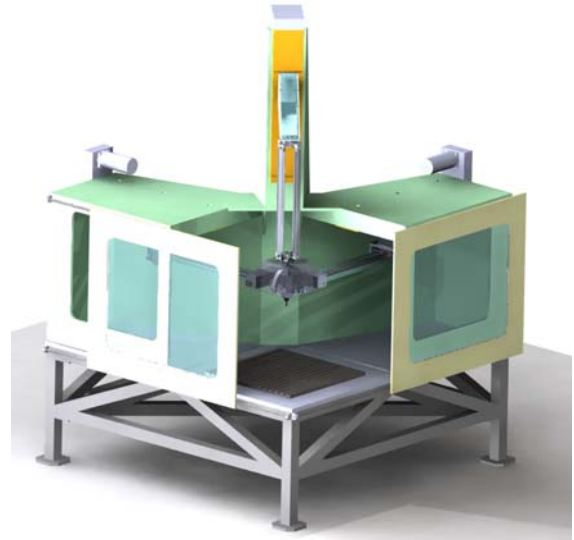

**Figure 1.** Orthoglide 5-axis

The two-dof spherical wrist that is implemented in Orthoglide 5-axis is derived from the Agile Eye, a three-dof spherical wrist [6]. Here, the two-dof spherical wrist is designed to obtain high stiffness [22]. A CAD model of the latter is shown in Figure 2

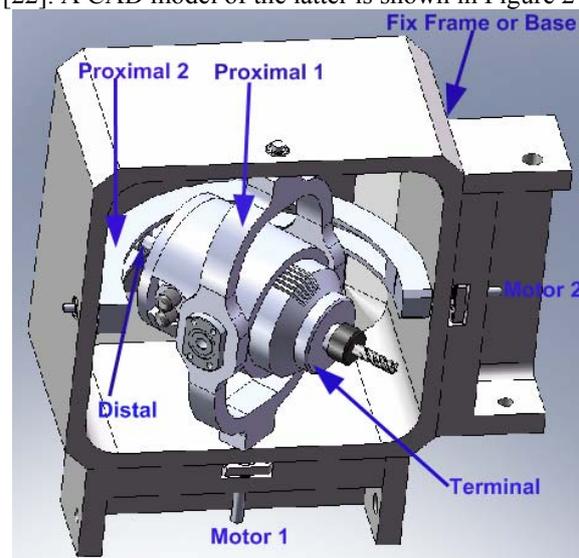

**Figure 2.** Spherical wrist of Orthoglide 5-axis

## 3 Spherical wrist kinematic model

## Reference frames and DH parameters

The spherical wrist of Orthoglide 5-axis is composed of a closed kinematic chain made up of five components: proximal-1, proximal-2, distal, terminal and the base. These five links are connected by means of revolute joints, of which axes intersect at the center of the mechanism. Besides, only the two revolute joints connected to the base of the wrist are actuated. The distal has an imaginary axis of rotation passing through the intersection point of other joint axis and perpendicular to the plane of proximal-2. A unit vector $e_i$ and a reference frame $R_i$ are associated to each joint with $Z_i$-axis and $e_i$ being coincident. The angle between $e_1$ and $e_2$ is denoted by $\alpha_0$ and the angle between $e_i$ and $e_{i+2}$ is denoted by $\alpha_i$, for i=1… 4.

Reference frame $R_1$ is defined in such a way that $Z_1$-axis coincides with $e_1$ and $e_2$ lies in the $X_1Z_1$-plane. Similarly $R_2$ has its $Z_2$-axis in the direction of $e_2$ and $e_1$ lies in the $X_2Z_2$-plane. Reference frame $R_i$ (i =3, 4, 5, 6) with $Z_i = e_i$ are defined by the rotation of frame $R_{i-2}$ and following the Denavit-Hartenberg conventions. DH-conventions for Orthoglide wrist mechanism are summarized as follow:

$Z_i$: axis of the i$^{th}$ joint;
$X_i$: common perpendicular to $Z_{i-2}$ and $Z_i$;
$Y_i$: respecting the right hand rule;
$a_i$: distance between $Z_i$ and $Z_{i+2}$;
$b_i$: distance between $X_i$ and $X_{i+2}$;
$\alpha_i$: angle between $Z_i$ and $Z_{i+2}$ about $X_{i+2}$;
$\theta_i$: angle between $X_i$ and $X_{i+2}$ about $Z_i$.

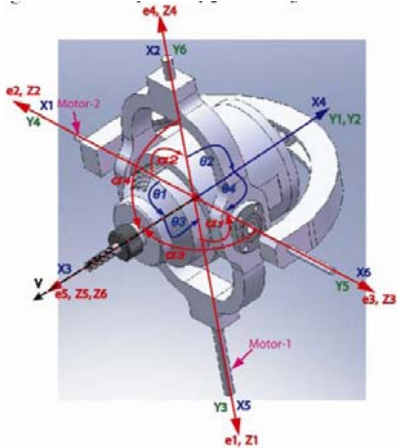

**Figure 3.** Orientations of reference frames according to DH-Conventions for Orthoglide Wrist

As all joints axes intersect at a common point, the origin of all the frames are the same i.e., $a_i = b_i = 0$. Figure 3 shows the orientations of reference frames attached to the Orthoglide wrist according to the DH convention, $\alpha_i$ being equal to $\pi/2$. In the home configuration, $\theta_1 = \theta_4 = -\pi/2$ and $\theta_2 = \theta_3 = +\pi/2$.

## Kinematics of Orthoglide wrist

The kinematics of the wrist of Orthoglide 5-axis is similar to the one of the Agile Eye introduced in [8].

However, the architecture is not the same as the distal and the proximal – 2 are different. Nevertheless, we can easily derive the kinematic and dynamic models of the wrist of Orthoglide 5-axis from [8]. Moreover, the influence of the external and/or machining forces on the dynamic performance is also considered.

Kinematic equations of the Orthoglide wrist are developed with the help of the reference frames defined above and the DH-parameters. It is noteworthy that the camera manipulator [8] and the Orthoglide wrist are analogous. However, subsequent modifications are made. Vector **v** depicts the orientation of the wrist end-effector (cutting tool etc), which is defined in reference frame $R_5$ by three angles $\beta_1$, $\beta_2$ & $\gamma$ illustrated in Figure 4:

- $\beta_1$ being the angle between $e_5$ and the projection of **v** to $e_3$-$e_5$ plane;
- $\beta_2$ being the angle between **v** and $e_3$-$e_5$ plane;
- $\gamma$ being the angle between **v** and $e_3$.

The expression of **v** in $R_5$ is then defined as,

$$[\mathbf{v}]_5 = [\sin\beta_2 \quad \sin\beta_1\cos\beta_2 \quad \cos\beta_1\cos\beta_2]^T$$

Since vectors **v** and $e_5$ coincide (Figure 3), i.e. **v** = $e_5$ hence $\beta_1 = \beta_2 = 0$ and $\gamma = \pi/2$.

The orientation of vector **v** is defined in reference frame $R_1$ by the pan ($\varphi_1$) and tilt ($\varphi_2$) angles as shown in Figure 4. With the definitions of $\varphi_1$ and $\varphi_2$, the components of **v** in $R_1$ are given by:

$$[\mathbf{v}]_1 = [\cos\varphi_1\cos\varphi_2 \quad \sin\varphi_1\cos\varphi_2 \quad \sin\varphi_2]^T$$

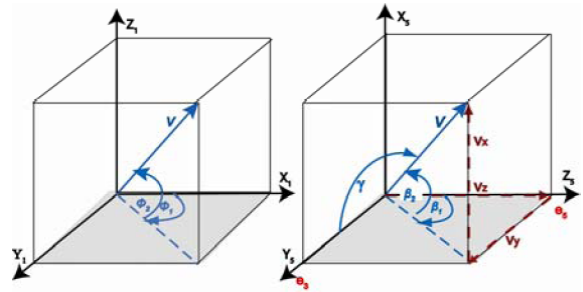

**Figure 4.** Definition of **v** in $R_1$ and $R_5$

Finally, the inverse kinematic problem of the wrist can be derived from **v**, $\alpha$, $\beta_1$, $\beta_2$, $\gamma$, $\varphi_1$ and $\varphi_2$. The relations for the joints variables ($\theta_1$, $\theta_2$, $\theta_3$, $\theta_4$) are summarized below [8] (in these relations, *c* and *s* stands for *cosine* and *sine* functions respectively).

$$\theta_1 = \tan^{-1}\left[2T/(1-T^2)\right]$$

where

$$T = \frac{-B + \sqrt{B^2 - 4AC}}{2A} \begin{cases} A = v_z c\alpha_1 + v_y s\alpha_1 - c\gamma \\ B = 2v_x s\alpha_1 \\ C = v_z c\alpha_1 - v_y c\alpha_1 - c\gamma \end{cases}$$

$$\theta_3 = \tan^{-1}\left(\frac{a\times e + b\times d}{a\times d - b\times e}\right)$$

where

$a = s\beta_2 \qquad b = s\alpha_3 c\beta_1 c\beta_2 - c\alpha_3 s\beta_1 c\beta_2$
$d = v_x c\theta_1 + v_y s\theta_1 \qquad e = -v_x s\theta_1 c\alpha_1 + v_y c\theta_1 c\alpha_1 + v_z c\alpha_1$

$$\theta_2 = \tan^{-1}\left(2T/\left(1-T_2^2\right)\right)$$

where

$$u_x = s\theta_1 s\alpha_1 c\alpha_3 + s\theta_1 c\theta_3 c\alpha_1 s\alpha_3 + c\theta_1 s\theta_3 s\alpha_3$$
$$u_y = -c\theta_1 s\alpha_1 c\alpha_3 - c\theta_1 c\theta_3 c\alpha_1 s\alpha_3 + s\theta_1 s\theta_3 s\alpha_3$$
$$u_z = c\alpha_1 c\alpha_3 - c\theta_3 c\alpha_1 c\alpha_3$$
$$A_2 = u_x s\alpha_0 c\alpha_2 + u_y s\alpha_2 + u_z c\alpha_0 c\alpha_2 - c\alpha_4$$
$$B_2 = 2\left(u_x c\alpha_0 s\alpha_2 - u_z s\alpha_0 s\alpha_2\right)$$
$$C_2 = u_x s\alpha_0 c\alpha_2 - u_y s\alpha_2 + u_z c\alpha_0 c\alpha_2 - c\alpha_4$$
$$T_2 = \left(-B_2 - \sqrt{B_2^2 - 4A_2 C_2}\right)/2A_2$$
$$\theta_4 = \tan^{-1}\left(s_4/c_4\right)$$

where

$$a_4 = -u_x\left(s\alpha_0 s\alpha_2 - s\theta_2 c\alpha_0 c\alpha_2\right) \quad, b_4 = -u_y c\theta_2 c\alpha_2$$
$$d_4 = -u_z\left(c\alpha_0 c\alpha_2 + s\theta_2 s\alpha_0 c\alpha_2\right)$$
$$s_4 = \frac{u_x c\theta_2 c\alpha_0 + u_y s\theta_2 - u_z c\theta_2 s\alpha_0}{s\alpha_4}, c_4 = \frac{a_4 + b_4 + d_4}{s\alpha_4}$$

Similarly, analytical relations of joints rates, i.e. joints velocities and acceleration can be obtained with the help of the previous relations of joint displacements and unit vectors $e_i$. Otherwise, numerical techniques, like finite difference method can be used to obtain joint velocities and accelerations.

## 4 Wrist dynamics

Dynamic analysis is of primary importance to investigate the forces and moments applied to the actuators to carry out a desired task or motion by the manipulator. Dynamic analysis of 2-DOF camera manipulator is presented in [8] where Newton-Euler approach is used. A similar methodology is also used here for Orthoglide wrist.

As a first step, wrist joints displacements are calculated from the kinematic model as discussed in the previous section. Velocities and accelerations of each component are obtained with the help of kinematic modeling, which has not been presented here because of space limitations. Also first and second derivatives of the unit vectors $e_1$, $e_2$, $e_3$, $e_4$ and $e_5$ are calculated. For the detailed relations of these calculations reader is referred to [8]. Finally a system of equilibrium equations, obtained from the free body diagrams of each wrist component, is used to get the relations of actuators torques. A flow chart of the torques calculations is given in Figure 5

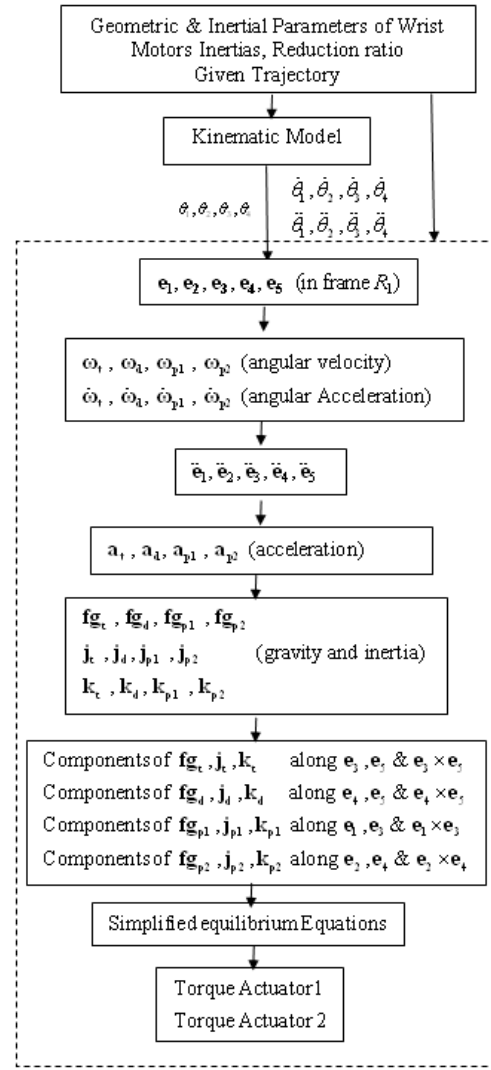

**Figure 5.** Flow chart of the dynamic model

**Equilibrium equations**

In this section, the methodology to obtain equilibrium equations of the Orthoglide wrist is presented. The following assumptions are made:
- friction forces are neglected;
- a spherical joint is assumed between the distal and the terminal link to have isostatic mechanism;
- a planar joint between the distal and the proximal -2.

With these assumptions, the free-body diagrams (FBD) of terminal, distal, proximal-1 and proximal-2 are drawn, as shown in Figure 6 to 9. Forces and moments acting on four moving wrist components are summarized below:
- two forces **A** and **B** and one moment **M** are exerted to the terminal;
- two force **C** and **D** are exerted to the distal;
- two forces **G** and **H** and two moments **N** and **P** are exerted to the proximal-1;
- two forces **E** and **F** and a moment **R** are exerted to the proximal-2.

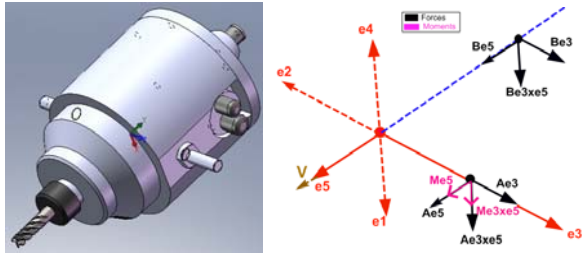

**Figure 6.** CAD model and FBD of terminal

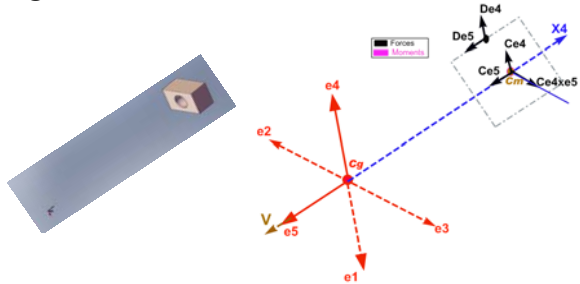

**Figure 7.** CAD model and FBD of distal

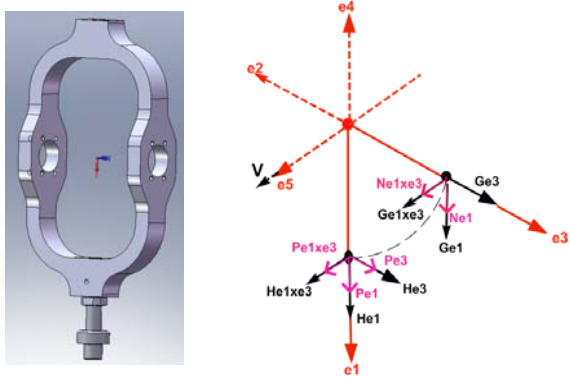

**Figure 8.** CAD model and FBD of proximl-1

These forces and moments are resolved into orthogonal components with respect to the attached unit vectors Equilibrium equations are written for each moving component (resulting 24 equations). For instance, set of equilibrium equations for the terminal (Figure 6) is given by:

$$\sum F_{e3} = A_{e3} + B_{e3} + F_{gte3} = J_{te3}$$

$$\sum F_{e5} = A_{e5} + B_{e5} + F_{gte5} = J_{te5}$$

$$\sum F_{e3 \times e5} = A_{e3 \times e5} + B_{e3 \times e5} + F_{gte3 \times e5} = J_{te3 \times e5}$$

$$\sum M_{e3} = -L_{te5} B_{e3 \times e5} - l_{te3 \times e5} F_{gte5} + l_{te5} F_{gte3 \times e5} = K_{te3}$$

$$\sum M_{e5} = M_{e5} - L_{te3} A_{e3 \times e5} - l_{te3} F_{gte3 \times e5} + l_{te3 \times e5} F_{gte3} = K_{te5}$$

$$\sum M_{e3 \times e5} = M_{e3 \times e5} + L_{te5} B_{e3} + L_{te3} A_{e5} + l_{te3} F_{gte5} - l_{te5} F_{gte3} = K_{te3 \times e5}$$

Where $F_{gte3}$, $F_{gte5}$, $F_{gte3xe5}$ are the gravity terms, $K_{gte3}$, $K_{gte5}$, $K_{gte3xe5}$ are the inertial terms (function of angular velocity, acceleration and inertia matrix) and $J_{gte3}$, $J_{gte5}$, $J_{gte3xe5}$ are the inertial forces (function of linear acceleration and mass) of the terminal along $e_3$, $e_5$ and $e_3xe_5$ directions respectively. Other variables used in these equations are defined in the next section.

Along with equilibrium equations, compatibility equations i.e., action-reaction equilibrium equations at terminal-distal, terminal-proximal-1 and distal-proximal-2 interaction points are also written (resulting 10 equations). The system of equations so obtained can be solved to obtain the torques experienced by the wrist actuators i.e. $P_{e1}$ and $R_{e2}$.

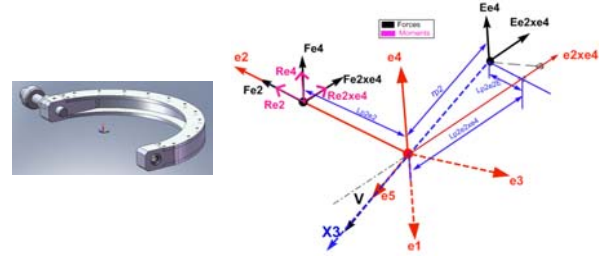

**Figure 9.** CAD model and FBD of proximl-2

## Wrist dynamic parameters

The input parameters of the dynamic model of the Orthoglide wrist are:
- mass of each component ($m_t$, $m_d$, $m_{p1}$, $m_{p2}$);
- distance between wrist geometric centre and centre of mass of each component ($l$);
- distance between wrist geometric centre and the point of application of force ($L$);
- inertia matrices of wrist component ($I_t$, $I_d$, $I_{p1}$, $I_{p2}$);
- inertia of actuators ($K_{m1}$, $K_{m2}$);
- reduction ratios of actuators ($\eta_{m1}$, $\eta_{m2}$).

The mass and inertial parameters are determined by means of a CAD software. The geometric parameters ($l$ or $L$) are determined from the drawings or CAD models of wrist components along with corresponding unit vectors.

## Effect of machining force on actuators torques

So far we have not taken into account the effect of the machining or cutting forces on the wrist dynamics. To cater for these forces we redraw the free body diagram of the terminal, with three components of cutting force $\mathbf{f}_c = \begin{bmatrix} F_{ce3} & F_{ce5} & F_{ce3 \times e5} \end{bmatrix}$

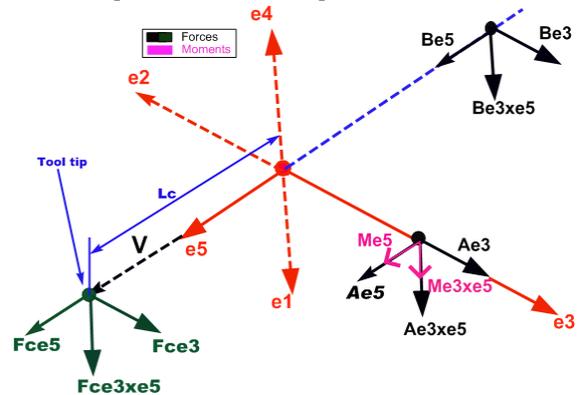

**Figure 10.** Terminal FBD with machining forces

The distance between the tool tip and geometric center of wrist is taken as $L_c$ (Figure 10). For this free body diagram, equilibrium equations are rewritten and new set of actuators equations are obtained.

### Motors parameters

In the preliminary design stage, FFA 20-80 harmonic drive motors are selected for both actuators. Motors specifications taken from the motor catalogue are given in Table 1 and theirs inertia is incorporated in the dynamic equations.

**Table 1.** Motors parameters

| FFA 20-80 Harmonic Drive | |
|---|---|
| Nominal speed | 2500 rpm |
| Maximum speed | 6500 rpm |
| Maximum torque at output shaft | 74 Nm |
| Continuous torque | 23 Nm |
| Motor moment of inertia | 0.00262 kg.m$^4$ |
| Motor power | 800 W |

### Trajectories generations

Two test trajectories are introduced for the Orthoglide 5-axis. Corresponding to these trajectories, inverse kinematic problem for the wrist is solved. The trajectories are defined as follows:

- Traj. I: semi-circular trajectory in vertical or YZ-plane defined by radius R and trajectory angles $\psi$ and $\delta$ (Figure 11)

$$\mathbf{v} = \begin{bmatrix} v_x \\ v_y \\ v_z \end{bmatrix} = \begin{bmatrix} 0 \\ -\sin\delta \\ -\cos\delta \end{bmatrix} \text{ where } \delta \text{ varies from } \pi/6 \text{ to }$$

$5\pi/6$ while $\psi$ varies from 0 to $\pi$.

- Traj. II: circular trajectory in horizontal or XY-plane defined by radius *R*, constant orientation angle $\gamma$ of vector **v** with Z-axis and angle $\delta$ (Figure 12)

$$\mathbf{v} = \begin{bmatrix} v_x \\ v_y \\ v_z \end{bmatrix} = \begin{bmatrix} \sin\gamma\cos\delta \\ \sin\gamma\sin\delta \\ -\cos\gamma \end{bmatrix}$$

where $\delta$ varies from 0 to $2\pi$

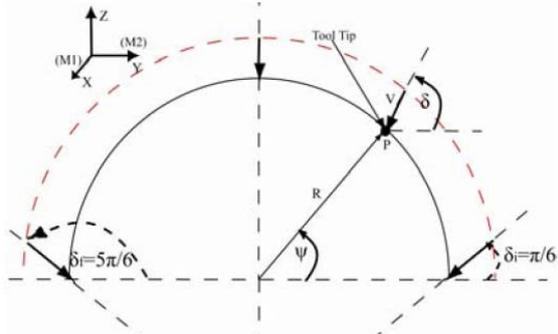

**Figure 11.** Orientation of vector v (Traj I)

## 5 Kinematic and dynamic analyses

The kinematics of the Orthoglide wrist is analysed for the two test trajectories. The velocity of the end-effector, located on point *P* (on the tip of the wrist tool), throughout the trajectory, is taken as constant i.e. $V_p$= 1 m/s and accordingly trajectory time is calculated with constant velocity.

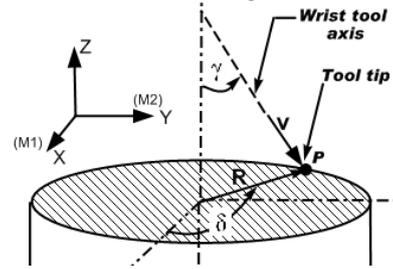

**Figure 12.** Orientation of vector **v** (Traj II)

Figure 13 shows the plots of position, velocity and acceleration of the joints for Traj. II with R= 0.25 m and $\gamma$= 45°. For the same trajectory, dynamic model is used to calculate the actuators torques, as shown in Figure 14.

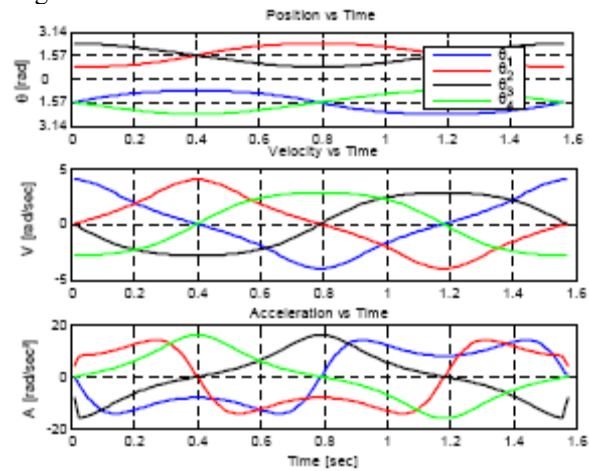

**Figure 13.** Position, velocity and acceleration of articulations (Traj II)

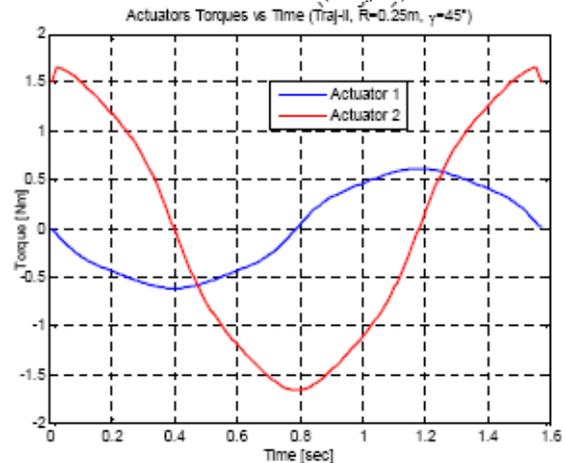

**Figure 14.** Actuators torques vs time (Traj. II)

Table 2 shows the maximum values of the kinematic and dynamic parameters for Traj. II for R= 0.05, 0.1, 0.15 and 0.25 m and for $\gamma$= 30˚, 45˚ and 60˚ where $T_1$ and $P_1$ (resp. $T_2$ and $P_2$) is the torque and the power of actuators 1 (resp. 2). Results show that the trajectory with R= 0.5 m and $\gamma$= 60˚ is more critical as compared to the other trajectories; it requires higher maximum velocities, accelerations and torques. Maximum values of actuators torques are also shown in Figure 15.

**Table 2.** Kinematics and dynamics peak values with no external forces (Traj II)

| γ [deg] | Radius [m] | Max Velocity [rad/s] | | | | Max Acceleration [rad/s2] | | | | Max Torque [Nm] | | Max Power [W] | |
|---|---|---|---|---|---|---|---|---|---|---|---|---|---|
| | | $\dot{\theta}_1$ | $\dot{\theta}_2$ | $\dot{\theta}_3$ | $\dot{\theta}_4$ | $\ddot{\theta}_1$ | $\ddot{\theta}_2$ | $\ddot{\theta}_3$ | $\ddot{\theta}_4$ | $T_1$ | $T_2$ | $P_1$ | $P_2$ |
| 30 | 0.250 | 2.31 | 2.31 | 2.00 | 2.00 | 7.29 | 7.29 | 9.21 | 9.21 | 0.68 | 1.22 | 0.64 | 1.20 |
| | 0.150 | 3.84 | 3.84 | 3.33 | 3.33 | 20.24 | 20.24 | 25.59 | 25.59 | 0.90 | 1.67 | 1.58 | 2.77 |
| | 0.100 | 5.77 | 5.77 | 5.00 | 5.00 | 45.54 | 45.54 | 57.58 | 57.59 | 1.35 | 2.55 | 3.89 | 6.43 |
| | 0.050 | 11.53 | 11.53 | 10.00 | 10.00 | 182.15 | 182.15 | 230.30 | 230.35 | 3.75 | 7.29 | 24.33 | 37.46 |
| 45 | 0.250 | 3.99 | 3.99 | 2.83 | 2.83 | 13.96 | 13.96 | 15.92 | 15.92 | 0.94 | 1.73 | 1.26 | 2.43 |
| | 0.150 | 6.65 | 6.65 | 4.71 | 4.71 | 38.78 | 38.78 | 44.22 | 44.22 | 1.23 | 2.37 | 3.86 | 5.97 |
| | 0.100 | 9.98 | 9.98 | 7.07 | 7.07 | 87.26 | 87.26 | 99.48 | 99.48 | 1.79 | 3.62 | 11.14 | 14.69 |
| | 0.050 | 19.96 | 19.96 | 14.14 | 14.14 | 349.03 | 349.03 | 397.94 | 397.94 | 5.58 | 10.38 | 80.14 | 92.25 |
| 60 | 0.250 | 6.90 | 6.90 | 3.46 | 3.46 | 34.05 | 34.05 | 27.36 | 27.36 | 1.11 | 2.16 | 3.17 | 5.22 |
| | 0.150 | 11.49 | 11.49 | 5.77 | 5.77 | 94.59 | 94.59 | 75.99 | 75.99 | 1.54 | 3.03 | 13.47 | 15.78 |
| | 0.100 | 17.24 | 17.24 | 8.66 | 8.66 | 212.82 | 212.82 | 170.98 | 170.98 | 3.31 | 4.72 | 44.17 | 45.04 |
| | 0.050 | 34.48 | 34.48 | 17.32 | 17.32 | 851.30 | 851.30 | 683.92 | 683.92 | 12.94 | 13.84 | 347.23 | 320.89 |

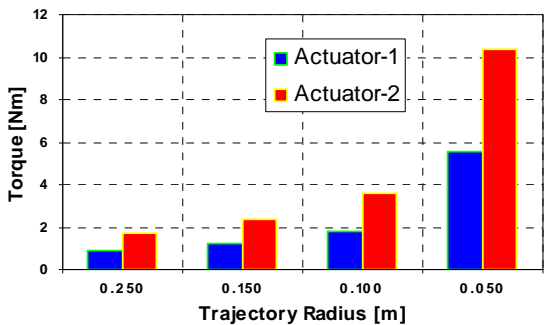

**Figure 15.** Actuators maximum torques vs trajectory radii (γ=45˚, Traj. II)

In order to analyse the effects of machining or cutting forces (see Figure 10), three equal components of **f_c** are assumed, i.e.,

$F_{ce3} = F_{ce3 \times e5} = F_{ce5} = F_c$ = constant

Actuators torques are calculated while considering the machining forces of different magnitudes and for trajectory radius of 0.15 m with γ= 45˚. Three values of machining forces moment arm $L_c$ are taken, i.e. $L_c$= 0.06, 0.11 and 0.15 m. Results are shown in Figure 16 to 18.

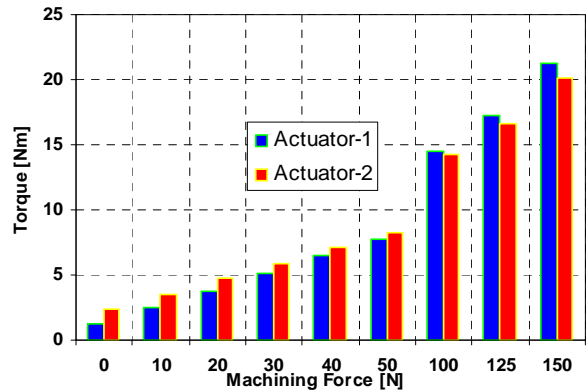

**Figure 17**. Actuators maximum torques with Fc (Lc= 0.11m, Traj. II)

Figure 18 shows that for the machining forces of 125 N and 150 N, actuators torques exceed the motors continuous torque (23 Nm) but still these are well below to the maximum motors torque (74 Nm). Hence for the given test trajectories considered motors can work for a range of machining forces. These results also represent the considerable influence of the length of the moment arm $L_c$ of the machining forces (or the tool length) on the actuators torques.

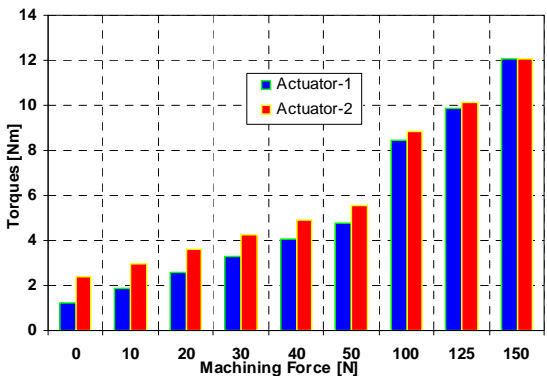

**Figure 16.** Actuators maximum torques with Fc (Lc= 0.06m, Traj. II)

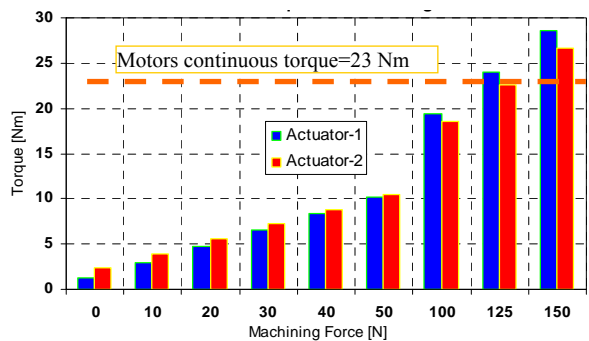

**Figure 18.** Actuators maximum torques with Fc (Lc= 0.15m, Traj. II)

## 6 Conclusions

This paper deals with the kinematics and dynamics of the spherical wrist of Orthoglide 5-axis. The kinematic and dynamic performances of the wrist were analyzed and its actuators primarily selection was proposed by means of several test trajectories. A methodology was introduced to evaluate the velocities, accelerations and torques required by the actuators. The influence of the machining forces as well as the tool length on the wrist actuators torques and powers was also studied. It turns out that the primarily selected motors with a continuous torque of 23 Nm and of power equal to 0.8 kW are suitable for the prototype of Orthoglide 5-axis. Finally, the following points will be taken into account in future works: (i) the friction between links has to be considered, (ii) planar joint between distal and proximal-2 should be analysed more precisely, (iii) weight of the machining tool should also be taken into consideration, and (iv) a larger range of trajectories should be analysed.